\documentclass[letterpaper, 10 pt, conference]{ieeeconf}  

\IEEEoverridecommandlockouts                              

\usepackage{graphics} 
\usepackage{graphicx}
\usepackage{epsfig} 
\usepackage{mathptmx} 
\usepackage{times} 
\usepackage{amsmath} 
\usepackage{amssymb}  
\usepackage[OT1, T1]{fontenc}
\usepackage[hyphens]{url}
\usepackage{booktabs}
\usepackage{cite}
\usepackage{multirow}
\usepackage{svg}
\usepackage{tcolorbox}
\usepackage[hidelinks]{hyperref}
\usepackage{xurl}

\title{\LARGE \bf
Act, Think or Abstain: Complexity-Aware Adaptive Inference for Vision-Language-Action Models
}

\author{Riccardo Andrea Izzo$^{\dag}$, Gianluca Bardaro$^{\dag}$, and Matteo Matteucci$^{\dag}$
\thanks{$^{\dag}$All authors are with the Department of Electronics, Information, and Bioengineering, Politecnico di Milano, Milan, Italy. E-mail:
        {\tt\small name.surname@polimi.it}}%
}

\begin{document}

\maketitle
\thispagestyle{empty}
\pagestyle{empty}

\begin{abstract}
Current research on Vision-Language-Action (VLA) models predominantly focuses on enhancing generalization through reasoning techniques. While effective, these improvements increase computational complexity and inference latency. Furthermore, these mechanisms are typically applied indiscriminately, wasting resources on trivial tasks while failing to provide the uncertainty estimation necessary to prevent catastrophic failure on out-of-distribution scenarios. Inspired by human cognition, we propose an adaptive framework that dynamically routes VLA execution based on the complexity of the perceived state. Our approach transforms the VLA's vision-language backbone into an active detection tool by projecting latent embeddings into a set of parametric and non-parametric estimators. This allows the system to execute known tasks immediately (\textit{Act}), reason about ambiguous scenarios (\textit{Think}), and preemptively halt execution when encountering physical or semantic anomalies (\textit{Abstain}). We find that a Gaussian Mixture Model fitted to fused vision-language embeddings provides the most reliable task-complexity signal, combining visual novelty with instruction context and cross-modal compatibility. Evaluated on the LIBERO and LIBERO-PRO benchmarks as well as on a real robot, our fused configuration achieves up to 87.5\% F1-score across two VLA backbones (SmolVLA and $\pi_0$), retains 83\% with as little as 5\% of training data, and surpasses state-of-the-art failure detectors.
\end{abstract}

\section{INTRODUCTION}
A unique trait of human intelligence is the ability to dynamically calibrate cognitive effort based on task demands. For routine tasks, we rely on fast and reactive decision-making, while for novel and ambiguous scenarios, we inherently analyze and reason about the task. When faced with a completely new task that goes beyond our abilities, we instinctively abstain to avoid damage. Nowadays, one promising direction for enhancing generalisation and adaptability in robotics is the use of Vision-Language-Action (VLA) models, end-to-end policies capable of grounding language and vision in the physical world. Despite their impressive capabilities, current VLAs lack this adaptive flexibility. Prior works have explored augmenting these with intermediate reasoning steps, such as Chain-of-Thought (CoT), significantly improving performance \cite{zawalski2025robotic, zhao2025cot, duan2025fast}. While effective, this push towards embodied reasoning comes at the non-negligible cost of increased computational complexity and inference latency, regardless of task difficulty. This results in the inefficient resource allocation for trivial tasks, and more critically, a failure to recognize when a task is completely out-of-distribution (OOD), leading to overconfident but catastrophic execution. In this paper, we advocate for a transition toward complexity-aware VLAs: a truly generalist policy should not act immediately, but consider the difficulty of the task before executing. We thus propose a framework that dynamically orchestrates VLA execution based on the complexity of the perceived state. Our approach leverages the VLA's pre-trained VLM backbone, transforming it from a passive feature extractor in latent space to an active complexity detector. This allows the system to execute known tasks with minimal latency overhead (\textit{Act}), perform additional reasoning when the environment or the task is ambiguous (\textit{Think}), or preemptively halt the execution when encountering heavy semantic or physical anomalies (\textit{Abstain}). Our pipeline scores the extracted embeddings using a set of parametric Gaussian Mixture Models (GMMs) and non-parametric k-Nearest Neighbour (kNN), with a simple Multi-Layer Perceptron (MLP) mapping the scores of vision, text, and fused features to the optimal execution strategy. Notably, our method is agnostic to both VLMs and different types of VLA architectures, and hence can be applied to diverse control policies, which we validate empirically by transferring the pipeline unchanged from SmolVLA to $\pi_0$. Our component analysis shows that vision and language provide complementary complexity cues: vision captures physical novelty, while instruction context exposes semantic ambiguity and cross-modal mismatch. Their fused representation yields the most reliable GMM score, whereas text alone remains insufficient and aggregation is not automatically beneficial.
Our key contributions include:
\begin{itemize}
    \item \textbf{Lightweight VLA-agnostic Complexity Detector:} A routing layer that repurposes the frozen VLM backbone of a VLA to detect task complexity, improving safety and efficiency without retraining the base policy.
    \item \textbf{Adaptive VLA Execution:} A three-way adaptive inference strategy that executes familiar tasks directly, invokes additional reasoning for ambiguous tasks, and abstains from tasks beyond the policy's capabilities, balancing computational requirements, responsiveness and generalization.
    \item \textbf{Multimodal Complexity Analysis:} We characterize how visual, textual, and fused VLA representations encode task complexity using parametric and non-parametric estimators, revealing complementary visual and linguistic cues and identifying parametric modelling of fused features as the most reliable approach.
\end{itemize}
\begin{figure*}[t]
    \centering
    \includegraphics[width=0.99\linewidth]{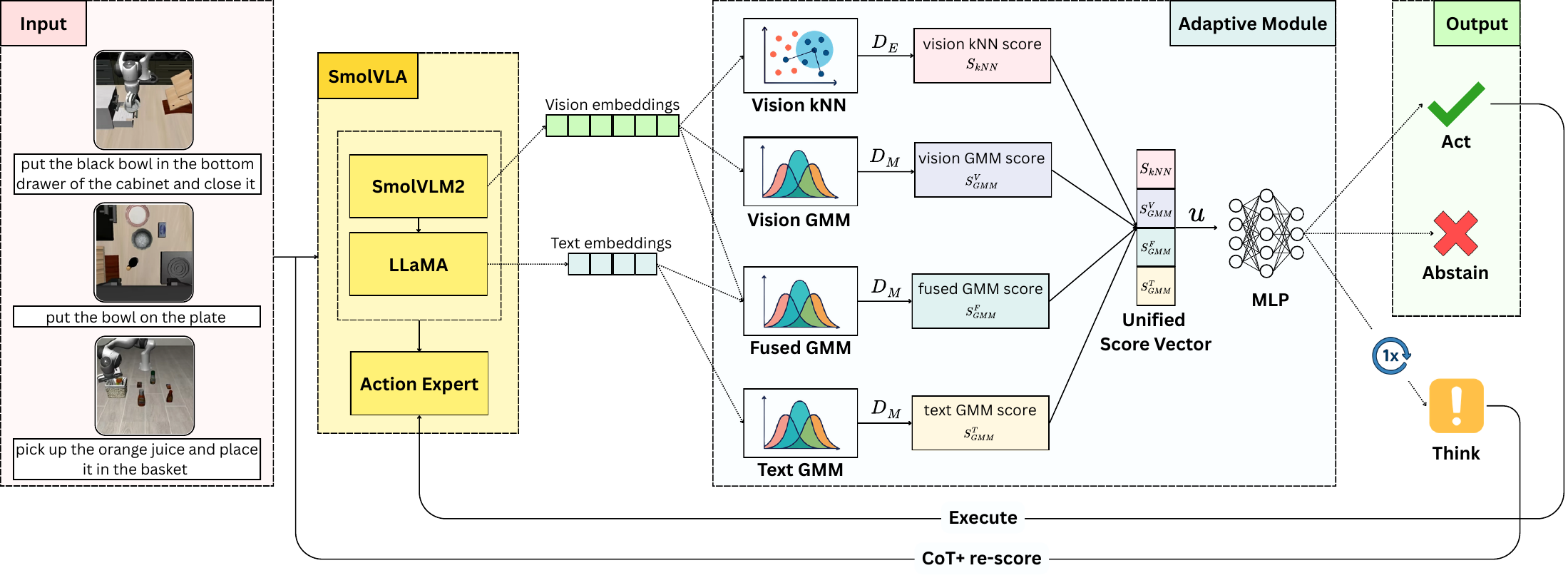}
    \caption{\textbf{Overview of our framework.} Given an image observation and a language instruction, the system extracts text and vision embeddings from the frozen VLA backbone (i.e., SmolVLA). These embeddings are then scored using a set of GMMs and kNN. The resulting scores are consolidated into a unified score vector and processed by a custom MLP to select the optimal execution strategy. High confidence triggers direct execution (\textit{Act}), while ambiguous scenarios result in additional reasoning (\textit{Think}). Tasks which are entirely out-of-distribution are preemptively halted (\textit{Abstain}). The \textit{Think} branch is invoked at most once per episode, after which the task resolves to either \textit{Act} or \textit{Abstain}.}
    \label{fig:framework}
\end{figure*}
By enabling VLAs to recognize the limits of their own capabilities, our framework provides a critical step toward deploying foundation models in safety-critical robotic environments. We open source code and models%
\footnote{\url{https://github.com/AIRLab-POLIMI/ActThinkAbstain}}.

\section{RELATED WORKS}
\subsection{Vision-Language-Action Models}
\label{sec:VLA}
Recent advancements in robot learning have increasingly relied on Vision-Language-Action (VLA) models, end-to-end systems that unify perception and language to generate robot actions, typically trained on massive multimodal datasets of text, vision, and robot trajectories \cite{o2024open}. Early works, such as RT-2 \cite{zitkovich2023rt} and OpenVLA \cite{kim2024openvla}, adapted standard VLMs to output actions as discretized text tokens, while subsequent research pivoted toward continuous action generation to overcome the precision limitations of discretization, with models like $\pi_0$ integrating flow-matching networks to produce high-frequency continuous trajectories \cite{black2024pi0}. Other works advocate for modular architectures that decouple high-level reasoning from low-level control, such as the dual-system design of GR00T \cite{bjorck2025gr00t}. Finally, to address the computational constraints of real-robot deployment, recent efforts such as SmolVLA develop VLAs on compact backbones while retaining competitive performance \cite{shukor2025smolvla}.

\subsection{Embodied Reasoning in VLAs}
\label{sec:embodied_reasoning}
Building upon standard VLA architectures, recent models enhance direct action prediction with intermediate reasoning steps. Approaches such as Embodied Chain-of-Thought (ECoT) \cite{zawalski2025robotic}, CoT-VLA \cite{zhao2025cot}, and InstructVLA \cite{yang2025instructvla} generate reasoning traces, sub-goals, or scene descriptions before execution, significantly improving success rates. However, running massive reasoning models at control frequency remains a computational bottleneck, and even latency-oriented variants such as Fast ECoT
\cite{duan2025fast} still enforce reasoning at every timestep, which is wasteful for trivial tasks. Most related to our motivation, OneTwoVLA \cite{lin2025onetwovla} unifies acting and reasoning within a single policy that adaptively decides when to
reason, and can even recover from erroneous executions in-context. These capabilities, however, are acquired by co-training the policy on synthesized reasoning data, tightly coupling the behavior to one specific model. In contrast, our framework treats reasoning as an explicit conditional fallback gated by a quantitative estimate of task complexity, requiring no modification of the underlying VLA. We view OneTwoVLA's in-context recovery as complementary, as it could be invoked within our \textit{Think} branch once a partially-OOD state is identified.

\subsection{Uncertainty Estimation and Failure Detection}
\label{sec:ood_detection}
Our approach addresses an OOD detection problem, a long-standing theme in imitation learning. Earlier approaches model different sources of uncertainty, using uncertainty from limited demonstration coverage to modulate robot compliance \cite{silverio2019uncertainty}, or aleatoric uncertainty to detect and resolve inconsistencies in the training data \cite{valletta2021imitation}. With the rise of generative policies, uncertainty has increasingly been derived from the policy's own objective: Diff-DAgger \cite{lee2025diff} uses the diffusion training loss to decide when to query an expert, while FAIL-Detect \cite{xu2025fail} flags runtime failures of diffusion-based policies without any failure data, combining a learned density score with conformal prediction. Closest to our setting, SAFE \cite{gusafe} trains a lightweight detector on a VLA's internal features to predict a scalar failure likelihood that transfers to unseen tasks. In contrast to these approaches, we do not score the action head or the denoising objective, but probe the frozen VLM backbone, disentangling visual novelty, linguistic ambiguity, and vision-language misalignment. Moreover, whereas prior detectors output a binary continue/abort decision, we resolve a three-way choice that isolates the recoverable, partially-OOD regime and routes it to reasoning rather than abstention, while remaining data-efficient and agnostic to the VLA architecture.

\section{METHOD}
Our approach leverages SmolVLA \cite{shukor2025smolvla} as the reference architecture, a VLA that integrates a pretrained SmolVLM-2 \cite{marafioti2025smolvlm} backbone with an action expert optimized via flow-matching \cite{lipman2023flow}. To enable the agent to be aware of task complexity and act accordingly, we propose a pipeline, illustrated in Figure~\ref{fig:framework}, that transforms the embeddings extracted from the VLM into a routing mechanism for adaptive inference: multimodal features are extracted from the backbone, scored against the In-Distribution (ID) manifold using parametric and non-parametric estimators, and finally mapped to our discrete set of strategies (i.e., Act, Think, Abstain).  

\subsection{Feature Extraction Pipeline}
\label{sec:feature_extraction}
We begin by extracting unimodal and multimodal features at inference time from the VLM backbone to capture linguistic ambiguity, visual novelty and vision-language misalignment. Given a batch of image observations and language instructions, the images are processed by the ViT encoder ($16\times16$ patches, 12 transformer layers), while language tokens are processed in parallel by the LLaMA \cite{grattafiori2024llama} text decoder. Let $B$ denote the batch size, $N_c$ the number of camera views, and $S_v, S_t$ the sequence lengths for visual and text inputs, respectively. The feature dimensions are $D_v = 768$ for the vision encoder and $D_t = 960$ for the text decoder. The feature extraction is defined as follows:

\noindent\textbf{Visual Features ($\mathbf{z}_{\mathrm{vis}}$):} Extracted from the last hidden states of the ViT encoder. To capture high-level semantic scene novelty before LLM projection, we perform spatial average pooling across the $S_v$ patches per camera, followed by mean pooling across all $N_c$ views to obtain $\mathbf{z}_{\mathrm{vis}} \in \mathbb{R}^{B \times D_{v}}$.

\noindent\textbf{Text Features ($\mathbf{z}_{\mathrm{text}}$):} Extracted from the LLaMA decoder's last hidden layer. We forward the language tokens without visual conditioning, treating the model as a pure text encoder so that the embeddings reflect only linguistic uncertainty rather than grounded scene information. We apply masked mean pooling across the sequence $S_t$, obtaining $\mathbf{z}_{\mathrm{text}} \in \mathbb{R}^{B \times D_{t}}$.

\noindent\textbf{Fused Features ($\mathbf{z}_{\mathrm{fused}}$):} To quantify vision-language mismatch, we implement a late fusion strategy. Both $\mathbf{z}_{\mathrm{vis}}$ and $\mathbf{z}_{\mathrm{text}}$ are normalized with $L_2$ and concatenated to form a joint representation $\mathbf{z}_{\mathrm{fused}} = [\bar{\mathbf{z}}_{\mathrm{vis}} \parallel \bar{\mathbf{z}}_{\mathrm{text}}] \in \mathbb{R}^{B \times (D_{v} + D_{t})}$.

\subsection{Distribution Fitting and OOD Scoring}
\label{sec:distribution_fitting}
Direct density estimation on the feature vectors $\mathbf{z}$ is often computationally expensive due to the curse of dimensionality. To mitigate this, we first apply Principal Component Analysis (PCA) to project the features into a lower-dimensional space $D' = 64$, preserving 95\% of the maximum variance and filtering out noise. The resulting reduced features, denoted as $\mathbf{z}'$, serve as input for our scoring modules.

\noindent\textbf{Gaussian Mixture Model (GMM)}: To account for the multimodal nature of robotic task clusters, we model the training distribution of the latent features as a GMM with $K$ components \cite{reynolds2009gaussian}. To quantify the novelty of a sample $\mathbf{z}'$, we compute the Mahalanobis distance \cite{de2000mahalanobis} to each Gaussian component $k$ in the mixture:
\begin{equation}
D_{M}(\mathbf{z}', \boldsymbol{\mu}_k, \boldsymbol{\Sigma}_k) = \sqrt{(\mathbf{z}' - \boldsymbol{\mu}_k)^\top \boldsymbol{\Sigma}_k^{-1} (\mathbf{z}' - \boldsymbol{\mu}_k)}
\end{equation}
where $\boldsymbol{\mu}_k$ and $\boldsymbol{\Sigma}_k$ are the sample mean and the covariance of the $k$-th component. Since scarce data may make $\boldsymbol{\Sigma}_{k}$ singular or ill-conditioned, we ensure invertibility through the Ledoit-Wolf shrinkage estimator \cite{ledoit2004well} with coefficient $\rho=0.01$. We finally define the GMM score, $S_{GMM}$, as the Mahalanobis distance $D_{M}$ to the closest Gaussian component $k$:
\begin{equation}
S_{GMM}(\mathbf{z}') = \min_{k} D_{M}(\mathbf{z}', \boldsymbol{\mu}_k, \boldsymbol{\Sigma}_k)
\end{equation}

\noindent\textbf{k-Nearest Neighbours (kNN)}: As a non-parametric alternative, we compute local density without assuming a global distribution. The kNN score is defined as the Euclidean distance to the nearest neighbour in the training set ${X}'_{train}$:
\begin{equation}
S_{kNN}(\mathbf{z}') = \min_{\mathbf{x}'_i \in {X}'_{train}} \|\mathbf{z}' - \mathbf{x}'_i\|_2
\end{equation}

We specifically utilize 1-NN to maximize sensitivity to subtle anomalies, ensuring that isolated and uncommon states remain clearly detectable rather than being smoothed out by neighbour averaging. The kNN score provides local sensitivity to outliers, while the GMM yields a probabilistic estimator that captures the global structure of complex task distributions. The two estimators also differ in deployment cost: the kNN search scales with the size of the reference set, whereas the fitted GMM is parametric, scoring each sample by computing $K$ Mahalanobis distances in the projected feature space of dimension $D'$, at a cost independent of the training set size. Rather than assuming their combination is beneficial, we treat the choice of estimator and modality as an empirical question, investigated in Section~\ref{sec:pipeline_effectiveness}.

\subsection{Score Aggregation}
\label{sec:scores}
The scores derived from visual, text, and fused representations provide a detailed view of uncertainty regarding the task complexity. However, as the magnitude and the ranges of these scores are linked to the statistical variance of the in-distribution training set, a direct approach based on thresholds is insufficient for effective decision making. We therefore propose to learn a function that maps these scores to a discrete system strategy. We first consolidate these scores into a unified vector $\mathbf{u} \in \mathbb{R}^4$:
\begin{equation}
\label{eq:score_vector}
\mathbf{u} = [S_{GMM}^{V}, S_{GMM}^{L}, S_{GMM}^{F}, S_{kNN}]^\top
\end{equation}
Equation \eqref{eq:score_vector} defines the general score vector, whose components are evaluated in Section~\ref{sec:pipeline_effectiveness}.
Notably, for the kNN estimator, we exclusively utilize visual features: text features in the training set exhibit high structural redundancy (e.g., repetitive task instructions), which collapses the local density in kNN, leading to extreme sensitivity to minor variations. Given the non-linear relationship between the computed scores and the actual success on the robot, we employ a MLP to predict the optimal execution strategy. The vector $\mathbf{u}$ is processed through Batch Normalization (BN) followed by two hidden layers with ReLU activations $\sigma$:
\begin{equation}
\mathbf{y} = \text{softmax}(W_3 \cdot \sigma(W_2 \cdot \sigma(W_1 \cdot \text{BN}(\mathbf{u}) + b_1) + b_2) + b_3)
\end{equation}
where $\mathbf{y}$ is the probability distribution over three distinct operational states. The final policy is determined as the $argmax$ of the output vector $\mathbf{y}$:
\begin{equation}
\text{Action} = \begin{cases}
\text{Act (ID)} & \text{if } \text{argmax}(\mathbf{y}) = 0 \\
\text{Think (Partially OOD)} & \text{if } \text{argmax}(\mathbf{y}) = 1 \\
\text{Abstain (OOD)} & \text{if } \text{argmax}(\mathbf{y}) = 2
\end{cases}
\label{eq:policy}
\end{equation}
The MLP maps the scores to one of three strategic outcomes, balancing efficiency and safety. In the \textit{Act} option, the task is recognized to be within the training distribution with high confidence, and the robot proceeds with immediate execution using the base VLA policy. Conversely, if the system selects \textit{Think}, it has detected a degree of semantic or visual ambiguity. This option pauses the execution to engage the VLM backbone in additional reasoning, extracting scene cues (e.g., object pose, object relations) and inferring subgoals from the current task instruction. As illustrated in Figure~\ref{fig:think_prompt}, this additional information is appended to the input text prompt, effectively grounding the VLM. We treat \textit{Think} as a transitional state rather than a terminal one: the enriched prompt changes the textual component of the fused representation, which is recomputed, re-scored by the fused GMM, and re-classified through Eq.~\ref{eq:policy}. The system proceeds to \textit{Act} only if the updated representation is classified as ID; a second \textit{Think} prediction or an \textit{Abstain} prediction conservatively resolves to \textit{Abstain}. This re-evaluation happens once per episode during the first timestep. For the vision-only GMM and kNN configurations, re-prompting leaves the visual embedding unchanged. Consequently, a transitional deployment of these configurations uses the fused GMM for post-\textit{Think} verification. The selected fused GMM configuration instead uses the same modality before and after reasoning. Finally, the \textit{Abstain} option provides a critical fallback for tasks that lie completely outside the model's capabilities.

\begin{figure}[t]
    \begin{tcolorbox}[
        colback=black!3,
        colframe=black!65,
        colbacktitle=black!65,
        coltitle=white,
        fonttitle=\footnotesize,
        boxrule=0.5pt,
        arc=1.5pt,
        left=4pt, right=4pt, top=3pt, bottom=3pt,
        toptitle=2pt, bottomtitle=2pt
    ]
    \footnotesize\ttfamily
    You are the reasoning module of a robot.\\ The task is: \textit{``\{instruction\}''}\\[2pt]
    Look at the image and answer in this format:\\
    Objects: \textit{<[color, type] of relevant objects>}\\
    Target: \textit{<the object the task refers to>}\\
    Steps: \textit{<list of task subgoals>}\\[2pt]
    Be concrete and brief. Do not add anything else.
    \tcblower
    \footnotesize
    \textbf{Enriched instruction:} \textit{\texttt{\{instruction\}}}. Objects: \textit{\texttt{\{objects\}}}. Target: \textit{\texttt{\{target\}}}. Plan: \textit{\texttt{\{subgoals\}}}.
    \end{tcolorbox}
    \caption{\textbf{\textit{Think} prompt template.} The VLM is queried with the current observation and the structured prompt (top). The output is appended to the task instruction (bottom) and re-scored by the fused GMM.}
    \label{fig:think_prompt}
\end{figure}

\begin{figure}[t]
    \centering
    \includegraphics[width=0.8\linewidth]{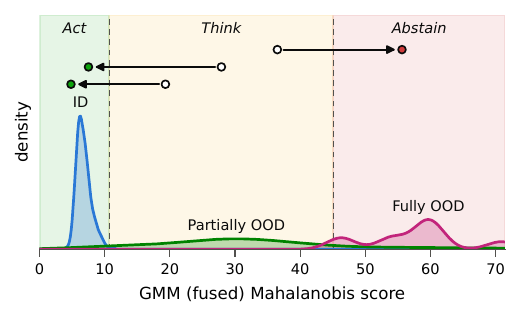}
    \caption{\textbf{Scores and learned routing decisions.} GMM (fused) scores
    for held-out ID, partially OOD, and fully OOD samples from $\mathcal{D}$,
    shown as unit-area Kernel Density Estimates (KDEs). Dashed lines show MLP
    decision boundaries, while hollow and filled markers with arrows illustrate
    post-\textit{Think} transitions to \textit{Act} or \textit{Abstain}.}
    \label{fig:fused_scores}
\end{figure}

\subsection{Training}
\label{sec:training}
\noindent\textbf{Dataset}: We construct the dataset $\mathcal{D}$ from rollouts collected on tasks drawn from LIBERO \cite{liu2023libero} and from the external Franka corpora (i.e., \textit{lerobot/nyu\_franka\_play\_dataset} and \textit{lerobot/cmu\_franka\_exploration\_dataset}). We use \textit{HuggingFaceVLA/smolvla\_libero}, a SmolVLA checkpoint fine-tuned on LIBERO. For each rollout $j$, we extract the visual, textual, and fused representations from the frozen VLM backbone at the first timestep, before execution, and record the final binary task outcome:
\begin{equation}
\mathcal{D} = \left\{\left(\mathbf{z}_{\mathrm{vis}}^{(j)}, \mathbf{z}_{\mathrm{text}}^{(j)}, \mathbf{z}_{\mathrm{fused}}^{(j)}, r^{(j)}\right)\right\}_{j=1}^{N},
\qquad r^{(j)} \in \{0,1\},
\end{equation}
where $r^{(j)}=1$ indicates successful task completion and $r^{(j)}=0$ indicates a failed rollout. The source corpus does not determine the label: LIBERO samples are not necessarily ID, nor are Franka samples necessarily OOD. Instead, the rollout outcome supervises the terminal decisions, with successful examples assigned to \textit{Act} and failed examples to \textit{Abstain}; the intermediate \textit{Think} class is synthesized by interpolating between them, as described below. LIBERO-PRO \cite{zhou2025libero}, comprising controlled object, position, semantic, task, and environment perturbations, is excluded from $\mathcal{D}$ and used exclusively for evaluation. Finally, we partition $\mathcal{D}$ into 50\% for detector fitting, 25\% for MLP training, and 25\% for validation to prevent episode-level leakage across splits.

\noindent\textbf{Detector Calibration}: As detailed in Section \ref{sec:distribution_fitting}, raw features are initially normalized and projected via PCA. We parameterised the GMMs using five random starts to mitigate the risk of local minima, selecting $K=3$ components by monitoring the macro F1-score on a validation set balanced between successful and failed rollouts. To this end, a single Gaussian cannot capture the multimodal structure of the task manifold, while larger values overfit local noise without accuracy gains. To support post-\textit{Think} re-evaluation, the reference set for the fused detector includes both embeddings from successful rollouts and CoT-augmented fused embeddings from resolved examples, all drawn from the partition of $\mathcal{D}$ related to detector fitting. Concurrently, a kNN index ($k=1$) is populated with the same projected features.

\noindent\textbf{MLP Training}: The MLP is trained to map the unified vector $\mathbf{u}$, or a selected subset of its components, to the three routing decisions defined in Equation~\ref{eq:policy}. We employ a lightweight architecture with two hidden layers of size 64 and 32, respectively. Training is conducted using a cross-entropy loss with a learning rate of $10^{-3}$ and early stopping on an internal held-out 15\% of the 25\% MLP training partition to prevent overfitting. A key challenge is supervising the intermediate \textit{Think} regime because curated examples are scarce and their interpretation depends on the deployed policy. Rather than assigning an entire benchmark to this class, we synthesize intermediate training features by interpolating between the ID and fully OOD reference distributions. Specifically, we employ a mixup strategy \cite{zhang2018mixup} using a Beta distribution:
\begin{equation}
\mathbf{z}_{think} = \lambda \mathbf{z}_{ID} + (1 - \lambda) \mathbf{z}_{OOD}, \quad \lambda \sim \text{Beta}(0.5, 0.5)
\end{equation}
Here, $\mathbf{z}_{ID}$ and $\mathbf{z}_{OOD}$ denote features sampled from the successful and failed reference subsets of $\mathcal{D}$, respectively.
This approach forces the MLP to learn a robust decision boundary for tasks neither fully familiar nor entirely novel.

\begin{figure}[t]
    \centering
    \includegraphics[width=0.9\linewidth]{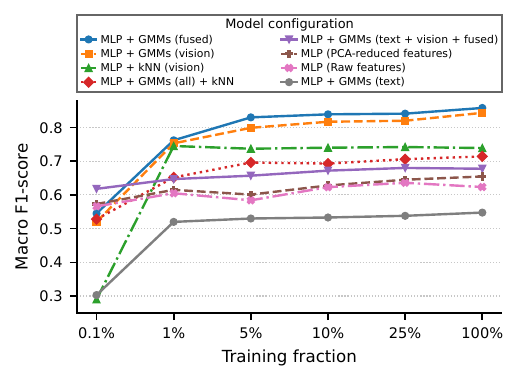}
    \caption{\textbf{Data scaling.} Macro F1 with increasing training data fractions.}
    \label{fig:performance}
\end{figure}

\noindent\textbf{Baseline MLP Training}: To quantify the efficacy of our system, we train an MLP directly on the concatenated embeddings $\mathbf{z}_{fused}$. To ensure a fair comparison, the baseline utilizes the same mixup strategy and a comparable lightweight architecture with two hidden layers of size 512 and 128, batch normalization, and dropout. We maintain the same learning rate of $10^{-3}$ with early stopping on the same internal held-out subset. We additionally train the same architecture on PCA-reduced fused features, isolating the contribution of the density estimators from that of dimensionality reduction.


\section{EXPERIMENTAL RESULTS}
We evaluate our framework on the LIBERO and LIBERO-PRO benchmarks, validating its ability to infer task complexity at inference time in scenarios where standard VLAs typically struggle due to distribution shifts. Specifically, we aim to address the following research questions (RQ):
\begin{itemize}
    \item \textbf{RQ1:} How does the amount of training data affect the ability of different scoring configurations to distinguish the three operating regimes?
    \item \textbf{RQ2:} What are the contributions of estimators, representation modality, and score aggregation to separating the three operating regimes across VLA backbones?
    \item \textbf{RQ3:} To what extent does adaptive routing prevent failures in OOD scenarios while preserving or improving task success in ID and partially OOD scenarios?
    \item \textbf{RQ4:} What is the trade-off between the computational overhead of our adaptive module and the gains in overall system reliability?
\end{itemize}

\begin{table}[t]
\centering
\caption{\textbf{Component analysis.} Precision (P), Recall (R), and Macro F1 (F1) on the three-class Act/Think/Abstain decision.}
\label{tab:component_analysis}
\resizebox{\columnwidth}{!}{%
\begin{tabular}{l ccc ccc}
\toprule
& \multicolumn{3}{c}{\textbf{SmolVLA}} & \multicolumn{3}{c}{\textbf{$\pi_0$}} \\
\cmidrule(lr){2-4} \cmidrule(lr){5-7}
\textbf{Method} & \textbf{P} & \textbf{R} & \textbf{F1} & \textbf{P} & \textbf{R} & \textbf{F1} \\
\midrule
MLP (Raw features)                             & 80.08 & 69.70 & 62.34 & 80.73 & 69.97 & 62.77 \\
MLP (PCA-reduced features)                     & 77.19 & 70.79 & 65.47 & 79.06 & 71.38 & 66.04 \\
\midrule
MLP + GMM (text + vision + fused)              & 72.57 & 70.67 & 67.76 & 74.04 & 71.45 & 68.42 \\
MLP + GMM (all) + kNN                          & 76.01 & 73.54 & 71.41 & 78.05 & 75.19 & 73.32 \\
\midrule
MLP + GMM (text)                               & 49.90 & 64.99 & 54.76 & 50.15 & 65.42 & 54.98 \\
MLP + kNN (vision)                             & 79.17 & 75.71 & 73.90 & 80.13 & 76.38 & 74.59 \\
MLP + GMM (vision)                             & 85.36 & 84.50 & 84.34 & 87.06 & 86.12 & 85.93 \\
MLP + GMM (fused)                              & \textbf{86.82} & \textbf{85.95} & \textbf{85.79} & \textbf{88.66} & \textbf{87.66} & \textbf{87.47} \\
\bottomrule
\end{tabular}%
}
\end{table}

\subsection{Data Scaling (RQ1)}
\label{sec:data_scaling}
To quantify the data required by our pipeline, we evaluate the framework across eight distinct configurations under an increasing number of data samples. We compare the baseline MLP on raw and reduced features against single-score GMM variants using text-only, vision-only, or fused features, a kNN variant, and ensemble configurations comprising both GMMs and kNN. Concretely, we subsample $\mathcal{D}$ at \{0.1\%, 1\%, 5\%, 10\%, 25\%\}, going from 64 to 15{,}885 samples. Figure \ref{fig:performance} reveals three key trends in Macro F1-score. First, the MLP directly trained on features remains almost insensitive to the size of the dataset, yielding stable but mediocre performance ($F1 \approx 0.60$) across all training fractions. This suggests that, without the density estimators, the model cannot effectively separate the manifold even when additional data is available. Second, the fused GMM exhibits the strongest scaling: it becomes the best configuration already at 1\% of the training data with an F1-score of 0.76, reaches 0.83 at 5\%, and then approaches saturation. The vision-only GMM follows closely, showing that visual structure provides a strong foundation while language supplies complementary task context. Finally, the broader multimodal GMM ensemble is the most robust configuration at 0.1\%, indicating that additional modalities can regularize extremely sparse training sets, but its advantage disappears as more data become available. Concurrently, the non-parametric kNN degrades significantly at 0.1\%, as local density estimates become critically sparse. Overall, the fused GMM reaches near top performance with as little as 5\% of the available training data, making it well suited to robotics applications with limited labeled data.

\begin{figure*}[t]
    \centering
    \includegraphics[width=1\linewidth]{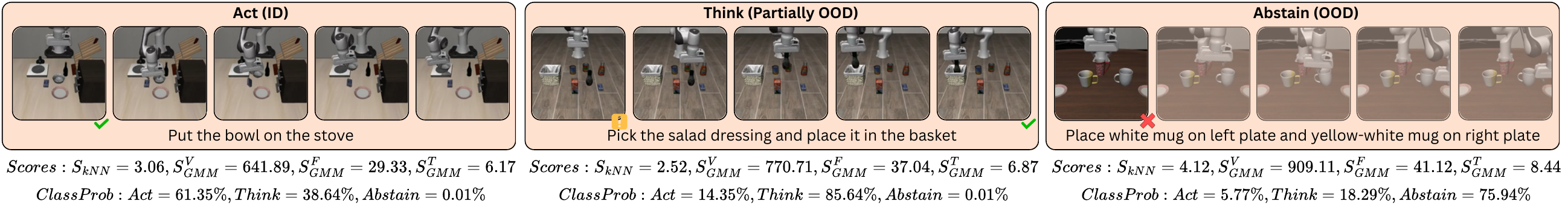}
    \caption{\textbf{Rollout examples on simulation.} Representative episodes from the LIBERO and LIBERO-PRO benchmarks: known tasks are executed with high confidence (left), ambiguous scenarios trigger additional reasoning leading to success (center), out-of-distribution tasks are preemptively halted (right).}
    \label{fig:simlation_example}
\end{figure*}

\subsection{Pipeline Effectiveness (RQ2)}
\label{sec:pipeline_effectiveness}
In Table \ref{tab:component_analysis} we evaluate the same configurations under the complete dataset $\mathcal{D}$, reporting Precision, Recall, and Macro F1-Score on its held-out validation partition averaged over three seeds to isolate the contribution of every component. To verify that the pipeline is VLA-agnostic, we repeat the component analysis on a $\pi_0$ \cite{black2024pi0} fine-tuned on LIBERO (i.e., \textit{lerobot/pi0\_libero\_base}), extracting embeddings from its VLM backbone and replicating the adaptive module training under the identical protocol of Section \ref{sec:training}.
Figure~\ref{fig:fused_scores} shows that the GMM (fused) score broadly orders the three operating regimes and supports post-\textit{Think} re-evaluation.

\noindent\textbf{Vision-Language Complementarity}: Our MLP + GMM (fused) configuration achieves the best result on SmolVLA, with a Macro F1-score of 85.79\%, followed closely by the vision-only GMM at 84.34\%. This shows that visual embeddings already provide a strong signal for separating ID, partially OOD, and fully OOD tasks, while instruction context and vision-language compatibility provide an additional, consistent gain when modeled jointly. The vision kNN remains competitive at 73.90\%, further confirming the importance of visual structure, although its local distance estimate is less effective than parametric density modeling.

\noindent\textbf{Baseline}: The baseline MLP trained directly on raw features obtains a Macro F1-score of 62.34\%. While it correctly identifies the \textit{Act} and \textit{Abstain} with high accuracy, it is overconfident on ambiguity, recalling only 10.5\% of \textit{Think} scenarios, potentially leading to unsafe execution in robotic tasks. Training the same MLP on PCA-reduced fused features yields only a marginal improvement (65.47\%), remaining far below the GMM-based configurations and indicating that the gain of our pipeline stems from explicit density estimation rather than dimensionality reduction. The text GMM nevertheless performs poorly, reaching 54.76\% Macro F1-score, due to repetitive instructions providing little evidence of physical scene novelty in isolation. Moreover, score-level aggregation of all GMMs and kNN reaches only 71.41\%, showing that simply adding estimators is not equivalent to learning a coherent joint representation. The benefit is specific to representation-level fusion: a single GMM fitted to the concatenated, normalized vision and text embeddings captures complementary cues without exposing the MLP to several redundant or weak scores.

\noindent\textbf{Robustness to Backbone Choice}: Repeating the component analysis on $\pi_0$ preserves the main ranking, with the fused GMM again achieving the best result (87.47\% Macro F1-score), narrowly ahead of the vision GMM (85.93\%), while the text variant remains the weakest. Absolute scores are uniformly slightly higher than their SmolVLA counterparts, consistent with the larger backbone yielding more informative embeddings. The consistent advantage of fusion indicates that complementary visual and linguistic complexity cues are present across VLM backbones rather than being specific to one architecture.

\subsection{Comparison with Failure Detection Baselines (RQ2)}
\label{sec:baseline_comparison}
In Table \ref{tab:baseline_comparison} we compare our best configuration from Section \ref{sec:pipeline_effectiveness} against two state-of-the-art failure detectors: logpZO~\cite{xu2025fail} and SAFE-MLP~\cite{gusafe}. Since both baselines produce a binary execute/abstain decision, we collapse our three-class output to the same binary space by mapping each \textit{Think} episode to its final resolved decision. For a fair comparison, we conduct the evaluation on the $\pi_0$ backbone, on which both detectors have been previously evaluated under the protocol of~\cite{gusafe}. We adopt the evaluation protocol of these works: treating abstention (i.e., failure) as the positive class, we report Balanced Accuracy (Bal-Acc), True Positive Rate (TPR), True Negative Rate (TNR), and the threshold-free ROC-AUC, using the abstention probability of our MLP as the continuous score. Unlike runtime monitors that score rollouts as they unfold, our framework decides before execution; we therefore compare decision quality on the same episodes rather than detection time. Our approach consistently outperforms logpZO and SAFE-MLP on both LIBERO and LIBERO-PRO across Balanced Accuracy, TPR, TNR, and ROC-AUC. The gains in both TPR and TNR show that it detects more failures while avoiding unnecessary abstentions, and the higher ROC-AUC indicates better overall separation between failure and non-failure cases.

\begin{table}[t]
\centering
\caption{\textbf{Failure detection baselines.} Binary execute/abstain decision with the $\pi_0$ backbone. Abstain is the positive class; Think episodes count as their final resolved decision.}
\label{tab:baseline_comparison}
\resizebox{\columnwidth}{!}{%
\begin{tabular}{l cccc cccc}
\toprule
& \multicolumn{4}{c}{\textbf{LIBERO}} & \multicolumn{4}{c}{\textbf{LIBERO-PRO}} \\
\cmidrule(lr){2-5} \cmidrule(lr){6-9}
\textbf{Method} & \textbf{Bal-Acc} & \textbf{TPR} & \textbf{TNR} & \textbf{ROC-AUC} & \textbf{Bal-Acc} & \textbf{TPR} & \textbf{TNR} & \textbf{ROC-AUC} \\
\midrule
logpZO~\cite{xu2025fail} & 65.58 & 55.41 & 75.75 & 77.67 & 75.42 & 75.46 & 75.37 & 80.91 \\
SAFE-MLP~\cite{gusafe} & 76.33 & 70.27 & 82.40 & 82.23 & 89.50 & 90.01 & 88.98 & 89.45 \\
\midrule
Ours & \textbf{83.66} & \textbf{79.73} & \textbf{87.59} & \textbf{92.23} & \textbf{93.58} & \textbf{95.36} & \textbf{91.80} & \textbf{94.12} \\
\bottomrule
\end{tabular}%
}
\end{table}

\begin{table*}[t]
\centering
\caption{\textbf{Simulation evaluation.} Performance on LIBERO and LIBERO-PRO with the SmolVLA and $\pi_0$ backbones.}
\label{tab:libero_results}
\resizebox{\textwidth}{!}{%
\begin{tabular}{ll cccccc c cccccc}
\toprule
& & \multicolumn{6}{c}{\textbf{SmolVLA}} & & \multicolumn{6}{c}{\textbf{$\pi_0$}} \\
\cmidrule(lr){3-8} \cmidrule(lr){10-15}
& & \multicolumn{2}{c}{\textbf{Baseline}} & \multicolumn{4}{c}{\textbf{Ours}} & & \multicolumn{2}{c}{\textbf{Baseline}} & \multicolumn{4}{c}{\textbf{Ours}} \\
\cmidrule(lr){3-4} \cmidrule(lr){5-8} \cmidrule(lr){10-11} \cmidrule(lr){12-15}
\textbf{Suite} & \textbf{Variant} & \textbf{SR (\%)} $\uparrow$ & \textbf{\boldmath$T_{inf}$ (s)} $\downarrow$ & \textbf{SR (\%)} $\uparrow$ & \textbf{PF} & \textbf{A (T) / Ab (T)} & \textbf{\boldmath$T_{inf}$ (s)} $\downarrow$ & & \textbf{SR (\%)} $\uparrow$ & \textbf{\boldmath$T_{inf}$ (s)} $\downarrow$ & \textbf{SR (\%)} $\uparrow$ & \textbf{PF} & \textbf{A (T) / Ab (T)} & \textbf{\boldmath$T_{inf}$ (s)} $\downarrow$ \\
\midrule
\multirow{6}{*}{\textbf{Goal}}
& base & 83.33 $\pm$ 15.28 & 49.71 & 86.67 $\pm$ 15.28 & 1 & 29 (1) / 1 (0) & 52.78 &  & 93.33 $\pm$ 5.77 & 58.99 & 96.67 $\pm$ 5.77 & 0 & 30 (1) / 0 (0) & 61.02 \\
& object & 60.00 $\pm$ 43.59 & 97.73 & 60.00 $\pm$ 43.59 & 10 & 19 (2) / 11 (3) & 61.42 &  & 93.33 $\pm$ 5.77 & 105.41 & 93.33 $\pm$ 5.77 & 1 & 29 (1) / 1 (0) & 104.23 \\
& position & 0.00 $\pm$ 0.00 & 163.46 & 0.00 $\pm$ 0.00 & 28 & 0 (0) / 30 (1) & 4.33 &  & 3.33 $\pm$ 5.77 & 188.63 & 10.00 $\pm$ 10.00 & 25 & 3 (0) / 27 (3) & 7.30 \\
& semantic & 90.00 $\pm$ 10.00 & 52.89 & 93.33 $\pm$ 11.55 & 2 & 28 (2) / 2 (1) & 53.80 &  & 86.67 $\pm$ 5.77 & 55.11 & 90.00 $\pm$ 10.00 & 1 & 28 (1) / 2 (1) & 46.97 \\
& task & 6.67 $\pm$ 11.55 & 152.24 & 6.67 $\pm$ 11.55 & 27 & 2 (1) / 28 (3) & 6.12 &  & 20.00 $\pm$ 10.00 & 125.52 & 26.67 $\pm$ 5.77 & 20 & 8 (1) / 22 (4) & 27.45 \\
& environment & 30.00 $\pm$ 10.00 & 109.21 & 30.00 $\pm$ 10.00 & 18 & 11 (4) / 19 (1) & 28.60 &  & 43.33 $\pm$ 5.77 & 83.12 & 53.33 $\pm$ 5.77 & 11 & 18 (3) / 12 (2) & 30.08 \\
\midrule
\multirow{6}{*}{\textbf{Object}}
& base & 90.00 $\pm$ 17.32 & 55.43 & 90.00 $\pm$ 17.32 & 1 & 28 (2) / 2 (1) & 54.82 &  & 83.33 $\pm$ 5.77 & 71.74 & 86.67 $\pm$ 5.77 & 1 & 29 (3) / 1 (0) & 49.26 \\
& object & 63.33 $\pm$ 37.86 & 91.12 & 66.67 $\pm$ 30.55 & 8 & 20 (4) / 10 (2) & 69.44 &  & 86.67 $\pm$ 11.55 & 59.93 & 93.33 $\pm$ 5.77 & 1 & 28 (2) / 2 (1) & 51.33 \\
& position & 0.00 $\pm$ 0.00 & 154.80 & 3.33 $\pm$ 5.77 & 28 & 1 (1) / 29 (5) & 8.58 &  & 3.33 $\pm$ 5.77 & 180.62 & 10.00 $\pm$ 10.00 & 25 & 5 (2) / 25 (1) & 8.21 \\
& semantic & 90.00 $\pm$ 10.00 & 55.14 & 96.67 $\pm$ 5.77 & 1 & 29 (4) / 1 (0) & 56.02 &  & 86.67 $\pm$ 5.77 & 60.11 & 90.00 $\pm$ 10.00 & 1 & 28 (3) / 2 (1) & 56.16 \\
& task & 0.00 $\pm$ 0.00 & 156.14 & 0.00 $\pm$ 0.00 & 29 & 0 (0) / 30 (1) & 3.56 &  & 0.00 $\pm$ 0.00 & 202.85 & 6.67 $\pm$ 5.77 & 27 & 2 (2) / 28 (3) & 10.39 \\
& environment & 16.67 $\pm$ 5.77 & 117.83 & 23.33 $\pm$ 5.77 & 19 & 11 (4) / 19 (3) & 38.19 &  & 30.00 $\pm$ 10.00 & 102.67 & 40.00 $\pm$ 0.00 & 15 & 15 (4) / 15 (2) & 44.43 \\
\midrule
\multirow{6}{*}{\textbf{Spatial}}
& base & 73.33 $\pm$ 25.17 & 51.83 & 80.00 $\pm$ 20.00 & 5 & 24 (2) / 6 (1) & 44.31 &  & 90.00 $\pm$ 0.00 & 48.21 & 93.33 $\pm$ 5.77 & 1 & 29 (1) / 1 (0) & 44.23 \\
& object & 70.00 $\pm$ 26.46 & 80.35 & 73.33 $\pm$ 25.17 & 7 & 22 (4) / 8 (2) & 64.34 &  & 93.33 $\pm$ 5.77 & 61.82 & 96.67 $\pm$ 5.77 & 0 & 30 (2) / 0 (0) & 57.41 \\
& position & 0.00 $\pm$ 0.00 & 157.42 & 0.00 $\pm$ 0.00 & 30 & 0 (0) / 30 (0) & 3.39 &  & 6.67 $\pm$ 11.55 & 171.45 & 10.00 $\pm$ 10.00 & 24 & 6 (3) / 24 (1) & 12.79 \\
& semantic & 73.33 $\pm$ 25.17 & 69.77 & 76.67 $\pm$ 20.82 & 6 & 24 (2) / 6 (3) & 60.97 &  & 93.33 $\pm$ 5.77 & 51.89 & 93.33 $\pm$ 5.77 & 1 & 28 (2) / 2 (1) & 52.21 \\
& task & 0.00 $\pm$ 0.00 & 155.92 & 3.33 $\pm$ 5.77 & 28 & 1 (1) / 29 (2) & 8.79 &  & 3.33 $\pm$ 5.77 & 171.88 & 23.33 $\pm$ 15.28 & 21 & 7 (3) / 23 (2) & 39.06 \\
& environment & 23.33 $\pm$ 5.77 & 98.33 & 33.33 $\pm$ 11.55 & 17 & 13 (3) / 17 (1) & 14.51 &  & 66.67 $\pm$ 5.77 & 85.16 & 66.67 $\pm$ 15.28 & 7 & 22 (2) / 8 (4) & 52.11 \\
\midrule
\multirow{6}{*}{\textbf{Long}}
& base & 53.33 $\pm$ 37.86 & 132.24 & 60.00 $\pm$ 30.00 & 10 & 18 (4) / 12 (2) & 80.91 &  & 73.33 $\pm$ 5.77 & 98.12 & 76.67 $\pm$ 15.28 & 3 & 24 (1) / 6 (3) & 94.12 \\
& object & 13.33 $\pm$ 23.09 & 149.40 & 16.67 $\pm$ 28.87 & 24 & 6 (3) / 24 (0) & 21.33 &  & 73.33 $\pm$ 5.77 & 96.12 & 76.67 $\pm$ 5.77 & 4 & 24 (1) / 6 (2) & 90.83 \\
& position & 0.00 $\pm$ 0.00 & 161.96 & 0.00 $\pm$ 0.00 & 29 & 0 (0) / 30 (1) & 4.78 &  & 3.33 $\pm$ 5.77 & 180.40 & 16.67 $\pm$ 5.77 & 23 & 7 (5) / 23 (1) & 40.99 \\
& semantic & 43.33 $\pm$ 35.12 & 136.38 & 50.00 $\pm$ 34.64 & 14 & 15 (4) / 15 (1) & 47.76 &  & 73.33 $\pm$ 5.77 & 58.23 & 86.67 $\pm$ 5.77 & 1 & 28 (2) / 2 (1) & 33.71 \\
& task & 0.00 $\pm$ 0.00 & 162.34 & 0.00 $\pm$ 0.00 & 28 & 0 (0) / 30 (2) & 5.26 &  & 0.00 $\pm$ 0.00 & 204.78 & 16.67 $\pm$ 5.77 & 23 & 7 (3) / 23 (0) & 14.88 \\
& environment & 13.33 $\pm$ 5.77 & 144.56 & 20.00 $\pm$ 10.00 & 20 & 8 (2) / 22 (2) & 29.17 &  & 23.33 $\pm$ 5.77 & 166.96 & 36.67 $\pm$ 5.77 & 15 & 14 (3) / 16 (1) & 38.90 \\
\bottomrule
\end{tabular}%
}
\end{table*}

\subsection{Simulation (RQ3, RQ4)}
We evaluate our best configuration (MLP + fused GMM) across both backbones, using the LIBERO base variants as the reference and LIBERO-PRO for distribution shifts. As defined in Section~\ref{sec:training}, recoverable shifts correspond to \textit{Think}, while shifts that remain unsuccessful correspond to \textit{Abstain}. Table~\ref{tab:libero_results} reports success rate (mean and standard deviation across three seeds), prevented failures (PF), and inference time. All timings, including Time-To-First-Action (TTFA), use the same NVIDIA RTX Quadro 6000. PF counts final abstentions on episodes that would otherwise fail and can therefore be lower than the total abstentions, with the difference corresponding to unnecessary abstentions on recoverable episodes. Final \textit{Act} (A) and \textit{Abstain} (Ab) decisions are reported with parentheses denoting those reached through \textit{Think}. Figure~\ref{fig:simlation_example} shows representative rollouts. On the base variants, more than 90\% of \textit{Goal} and \textit{Object} episodes on both backbones follow the \textit{Act} path with inference times comparable to the baselines. The \textit{Think} branch recovers episodes where the baseline fails, improving success by 6.67\% on the \textit{Spatial} and \textit{Long} suites. Across all 24 variants, routing never reduces success, adding 23 successes over 720 SmolVLA episodes and 48 with $\pi_0$, confirming that adaptivity comes at no cost in task performance. Under partial OOD shifts, including semantic and object variants, the pipeline increasingly selects \textit{Think} or \textit{Abstain} while preserving performance and avoiding wasted computation. For instance, on the \textit{Goal} object variant it matches baseline success while abstaining on 11 episodes, 10 of which are prevented failures, and reduces inference time from 98s to 61s. Under the severe \textit{position} and \textit{task} shifts, the SmolVLA baseline often runs for more than 150s before failing, whereas our pipeline abstains in 236 of 240 episodes and reduces inference time by 96\% on average. Overall, 641 of 682 abstentions (94\%) prevent failures. On the same severe shifts, $\pi_0$ abstains less frequently (195 of 240 episodes) while retaining a non-zero success rate, showing that the detector tracks the capability of the underlying policy rather than flagging anomalous inputs. Routing introduces a TTFA overhead, measured as the additional delay before the first action relative to execution without routing: 523 ms for \textit{Act} and 1.612 s for \textit{Think}, the latter including reasoning and subsequent re-evaluation, after which control resumes at the baseline frequency. Despite this overhead, average inference time remains lower because the \textit{Abstain} path ends OOD tasks early. Finally, rare OOD cases routed to \textit{Think} subsequently failed, revealing a conservative bias toward recovery.

\begin{figure}[t]
    \centering
    \includegraphics[width=0.9\linewidth]{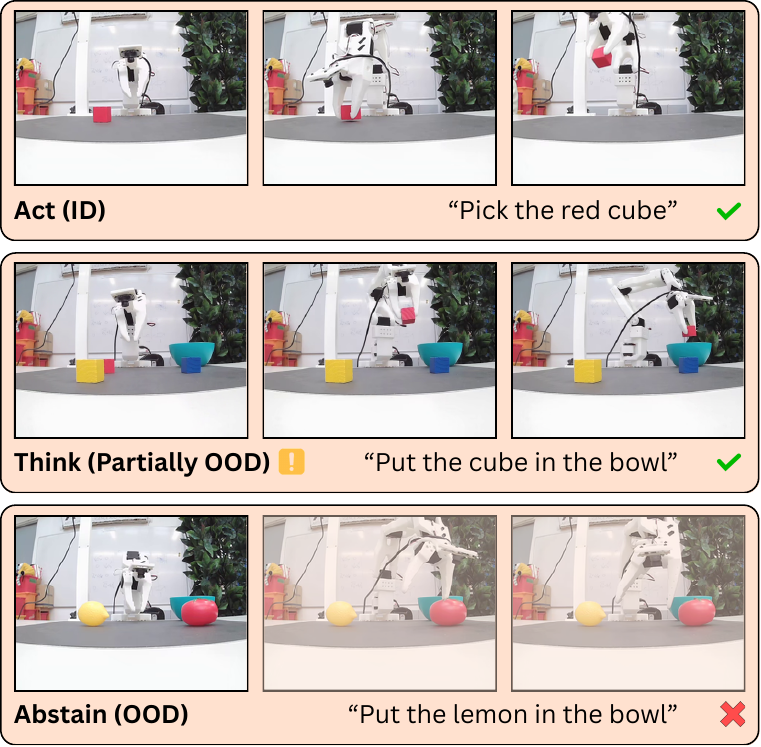}
    \caption{\textbf{Real-robot rollout examples.} SO-ARM 101 with SmolVLA performing tabletop manipulation tasks.}
    \label{fig:real_robot_example}
\end{figure}

\subsection{Real Robot (RQ3)}
We fine-tune SmolVLA on 100 trajectories collected on an SO-ARM 101, establishing the baseline for the \textit{Act} path. The detector and router are refit on these 100 rollouts following Section~\ref{sec:training}. We evaluate 30 tasks (10 per regime) over three trials, giving 90 rollouts per method. ID tasks comprise picking, pick-and-place, and stacking with the training objects. Partially OOD tasks keep those instructions but add unseen distractors. Fully OOD tasks reference objects absent from training and ambiguous instructions, where the base policy typically fails. Table~\ref{tab:real_robot} reports per-regime SR (mean $\pm$ std), PF, routing decisions, and inference time, with representative rollouts shown in Figure~\ref{fig:real_robot_example}. Our framework improves success in every regime while cutting mean inference time by 58.6\% (170.8s to 70.6s). On ID tasks all 30 rollouts execute, four through \textit{Think}, lifting success to 96.7\% at unchanged cost. Partial shifts split evenly between execution and abstention, and \textit{Think} adds 13.3 points. Under full OOD shifts routing abstains on 25 of 30 episodes and cuts time by 88\%, while \textit{Think} still recovers five executions, four of which succeed. Overall, 34 of 40 abstentions prevent failures, confirming that our results transfer to hardware.

\begin{table}[t]
\centering
\caption{\textbf{Real-robot evaluation.} SmolVLA on the SO-ARM 101.}
\label{tab:real_robot}
\resizebox{\columnwidth}{!}{%
\begin{tabular}{l cc cccc}
\toprule
& \multicolumn{2}{c}{\textbf{Baseline}} & \multicolumn{4}{c}{\textbf{Ours}} \\
\cmidrule(lr){2-3} \cmidrule(lr){4-7}
\textbf{Regime} & \textbf{SR (\%)} $\uparrow$ & \textbf{\boldmath$T_{inf}$ (s)} $\downarrow$ & \textbf{SR (\%)} $\uparrow$ & \textbf{PF} & \textbf{A (T) / Ab (T)} & \textbf{\boldmath$T_{inf}$ (s)} $\downarrow$ \\
\midrule
ID            & 90.0 $\pm$ 10.0 & 114.8 & \textbf{96.7} $\pm$ 5.8 & 0 & 30 (4) / 0 (0) & \textbf{108.6} \\
Partially OOD & 36.7 $\pm$ 5.8 & 190.3 & \textbf{50.0} $\pm$ 0.0 & 11 & 15 (11) / 15 (8) & \textbf{78.2} \\
Fully OOD     & 6.7 $\pm$ 5.8 & 207.3 & \textbf{13.3} $\pm$ 5.8 & 23 & 5 (5) / 25 (9) & \textbf{25.1} \\
\bottomrule
\end{tabular}%
}
\end{table}

\section{CONCLUSIONS}
We introduced an adaptive VLA framework that balances performance, efficiency, and safety by routing ID tasks to execution, partial shifts to reasoning, and fully OOD tasks to abstention. Across SmolVLA and $\pi_0$, a GMM over fused embeddings from the frozen VLM backbone provides the most reliable complexity signal and transfers without modifying the underlying policies. Results on LIBERO, LIBERO-PRO and on a real robot support complexity-aware inference as a path toward robust robotic foundation models.

Despite these results, some limitations remain. While our fused GMM performs convincingly, there is a gap in recovering partially OOD tasks that are classified as ID. This may be due to routing being formulated as a classification problem, which creates rigid boundaries at the edges of the distribution shifts. A solution would be to formulate routing as a continuous prediction problem, with adaptive thresholds optimized using task-success feedback. Moreover, our routing acts once per episode before execution and does not yet react to mid-execution perturbations, such as human interference. Re-evaluating task complexity online during rollout is a promising direction for closed-loop safety. Finally, to remove the reliance on known ID and OOD reference sets, we plan to investigate zero-shot adaptation using vision-language alignment strategies and token uncertainty.

\bibliographystyle{IEEEtran}
\bibliography{bibliography}

@inproceedings{o2024open,
  title={Open x-embodiment: Robotic learning datasets and rt-x models},
  author={O’Neill, Abby and Rehman, Abdul and Maddukuri, Abhiram and Gupta, Abhishek and Padalkar, Abhishek and Lee, Abraham and others},
  booktitle={2024 IEEE International Conference on Robotics and Automation (ICRA)},
  pages={6892--6903},
  year={2024},
  organization={IEEE}
}

@inproceedings{zitkovich2023rt,
  title={Rt-2: Vision-language-action models transfer web knowledge to robotic control},
  author={Zitkovich, Brianna and Yu, Tianhe and Xu, Sichun and Xu, Peng and Xiao, Ted and Xia, Fei and others},
  booktitle={Conference on Robot Learning},
  pages={2165--2183},
  year={2023},
  organization={PMLR}
}

@inproceedings{kim2024openvla,
  title = 	 {OpenVLA: An Open-Source Vision-Language-Action Model},
  author={Kim, Moo Jin and Pertsch, Karl and Karamcheti, Siddharth and Xiao, Ted and Balakrishna, Ashwin and Nair, Suraj and others},
  booktitle = 	 {Proceedings of The 8th Conference on Robot Learning},
  year = 	 {2024},
}

@article{black2024pi0,
  title={{$\pi_0$}: A Vision-Language-Action Flow Model for General Robot Control},
  author={Black, Kevin and Brown, Noah and Driess, Danny and Esmail, Adnan and Equi, Michael and Finn, Chelsea and others},
  journal={arXiv preprint arXiv:2410.24164},
  year={2024}
}

@article{shukor2025smolvla,
  title={Smolvla: A vision-language-action model for affordable and efficient robotics},
  author={Shukor, Mustafa and Aubakirova, Dana and Capuano, Francesco and Kooijmans, Pepijn and Palma, Steven and Zouitine, Adil and others},
  journal={arXiv preprint arXiv:2506.01844},
  year={2025}
}

@article{marafioti2025smolvlm,
  title={Smolvlm: Redefining small and efficient multimodal models},
  author={Marafioti, Andr{\'e}s and Zohar, Orr and Farr{\'e}, Miquel and Noyan, Merve and Bakouch, Elie and Cuenca, Pedro and others},
  journal={arXiv preprint arXiv:2504.05299},
  year={2025}
}

@inproceedings{lipman2023flow,
  title={Flow Matching for Generative Modeling},
  author={Lipman, Yaron and Chen, Ricky TQ and Ben-Hamu, Heli and Nickel, Maximilian and Le, Matthew},
  booktitle={The Eleventh International Conference on Learning Representations},
  year={2023}
}

@article{grattafiori2024llama,
  title={The llama 3 herd of models},
  author={Grattafiori, Aaron and Dubey, Abhimanyu and Jauhri, Abhinav and Pandey, Abhinav and Kadian, Abhishek and Al-Dahle, Ahmad and others},
  journal={arXiv preprint arXiv:2407.21783},
  year={2024}
}

@article{bjorck2025gr00t,
  title={Gr00t n1: An open foundation model for generalist humanoid robots},
  author={Bjorck, Johan and Casta{\~n}eda, Fernando and Cherniadev, Nikita and Da, Xingye and Ding, Runyu and Fan, Linxi and others},
  journal={arXiv preprint arXiv:2503.14734},
  year={2025}
}

@article{liu2023libero,
  title={Libero: Benchmarking knowledge transfer for lifelong robot learning},
  author={Liu, Bo and Zhu, Yifeng and Gao, Chongkai and Feng, Yihao and Liu, Qiang and Zhu, Yuke and others},
  journal={Advances in Neural Information Processing Systems},
  volume={36},
  pages={44776--44791},
  year={2023}
}

@article{zhou2025libero,
  title={LIBERO-PRO: Towards Robust and Fair Evaluation of Vision-Language-Action Models Beyond Memorization},
  author={Zhou, Xueyang and Xu, Yangming and Tie, Guiyao and Chen, Yongchao and Zhang, Guowen and Chu, Duanfeng and others},
  journal={arXiv preprint arXiv:2510.03827},
  year={2025}
}

@article{yang2025instructvla,
  title={Instructvla: Vision-language-action instruction tuning from understanding to manipulation},
  author={Yang, Shuai and Li, Hao and Chen, Yilun and Wang, Bin and Tian, Yang and Wang, Tai and others},
  journal={arXiv preprint arXiv:2507.17520},
  year={2025}
}

@inproceedings{zawalski2025robotic,
  title={Robotic Control via Embodied Chain-of-Thought Reasoning},
  author={Zawalski, Micha{\l} and Chen, William and Pertsch, Karl and Mees, Oier and Finn, Chelsea and Levine, Sergey},
  booktitle={Conference on Robot Learning},
  pages={3157--3181},
  year={2024},
  organization={PMLR}
}

@inproceedings{zhao2025cot,
  title={Cot-vla: Visual chain-of-thought reasoning for vision-language-action models},
  author={Zhao, Qingqing and Lu, Yao and Kim, Moo Jin and Fu, Zipeng and Zhang, Zhuoyang and Wu, Yecheng and others},
  booktitle={Proceedings of the Computer Vision and Pattern Recognition Conference},
  pages={1702--1713},
  year={2025}
}

@article{duan2025fast,
  title={Fast ecot: Efficient embodied chain-of-thought via thoughts reuse},
  author={Duan, Zhekai and Zhang, Yuan and Geng, Shikai and Liu, Gaowen and Boedecker, Joschka and Lu, Chris Xiaoxuan},
  journal={arXiv preprint arXiv:2506.07639},
  year={2025}
}

@article{lin2025onetwovla,
  title={Onetwovla: A unified vision-language-action model with adaptive reasoning},
  author={Lin, Fanqi and Nai, Ruiqian and Hu, Yingdong and You, Jiacheng and Zhao, Junming and Gao, Yang},
  journal={arXiv preprint arXiv:2505.11917},
  year={2025}
}

@inproceedings{gusafe,
  title={SAFE: Multitask Failure Detection for Vision-Language-Action Models},
  author={Gu, Qiao and Ju, Yuanliang and Sun, Shengxiang and Gilitschenski, Igor and Nishimura, Haruki and Itkina, Masha and others},
  booktitle={The Thirty-ninth Annual Conference on Neural Information Processing Systems},
  year={2025}
}

@inproceedings{silverio2019uncertainty,
  title={Uncertainty-aware imitation learning using kernelized movement primitives},
  author={Silv{\'e}rio, Jo{\~a}o and Huang, Yanlong and Abu-Dakka, Fares J and Rozo, Leonel and Caldwell, Darwin G},
  booktitle={2019 IEEE/RSJ International Conference on Intelligent Robots and Systems (IROS)},
  pages={90--97},
  year={2019},
  organization={IEEE}
}

@inproceedings{valletta2021imitation,
  title={Imitation learning with inconsistent demonstrations through uncertainty-based data manipulation},
  author={Valletta, Peter and P{\'e}rez-Dattari, Rodrigo and Kober, Jens},
  booktitle={2021 IEEE International Conference on Robotics and Automation (ICRA)},
  pages={3655--3661},
  year={2021},
  organization={IEEE}
}

@inproceedings{lee2025diff,
  title={Diff-dagger: Uncertainty estimation with diffusion policy for robotic manipulation},
  author={Lee, Sung-Wook and Kang, Xuhui and Kuo, Yen-Ling},
  booktitle={2025 IEEE International Conference on Robotics and Automation (ICRA)},
  pages={4845--4852},
  year={2025},
  organization={IEEE}
}

@inproceedings{xu2025fail,
  title={Can We Detect Failures Without Failure Data? {U}ncertainty-Aware Runtime Failure Detection for Imitation Learning Policies},
  author={Xu, Chen and Nguyen, Tony Khuong and Dixon, Emma and Rodriguez, Christopher and Miller, Patrick and Lee, Robert and others},
  booktitle={Proceedings of Robotics: Science and Systems (RSS)},
  year={2025},
  doi={10.15607/RSS.2025.XXI.073}
}

@article{reynolds2009gaussian,
  title={Gaussian mixture models.},
  author={Reynolds, Douglas A and others},
  journal={Encyclopedia of Biometrics},
  pages={659--663},
  year={2009},
  publisher={Springer}
}

@article{de2000mahalanobis,
  title={The mahalanobis distance},
  author={De Maesschalck, Roy and Jouan-Rimbaud, Delphine and Massart, D{\'e}sir{\'e} L},
  journal={Chemometrics and intelligent laboratory systems},
  volume={50},
  number={1},
  pages={1--18},
  year={2000},
  publisher={Elsevier}
}

@article{ledoit2004well,
  title={A well-conditioned estimator for large-dimensional covariance matrices},
  author={Ledoit, Olivier and Wolf, Michael},
  journal={Journal of multivariate analysis},
  volume={88},
  number={2},
  pages={365--411},
  year={2004},
  publisher={Elsevier}
}

@inproceedings{zhang2018mixup,
  title={mixup: Beyond Empirical Risk Minimization},
  author={Zhang, Hongyi and Cisse, Moustapha and Dauphin, Yann N and Lopez-Paz, David},
  booktitle={International Conference on Learning Representations},
  year={2018}
}

\addtolength{\textheight}{-12cm}   

\end{document}